# ON A CEPSTRUM-BASED SPEECH DETECTOR ROBUST TO WHITE NOISE

*Sergei Skorik and Frédéric Berthommier*

Institut de la Communication Parlée (ICP/INPG)
46 Av Félix Viallet, 38031 Grenoble, Cedex 1, France
e-mail: seryi@fcmail.com,  bertho@icp.inpg.fr

## ABSTRACT

We study effects of additive white noise on the cepstral representation of speech signals. Distribution of each individual cepstrum coefficient of speech is shown to depend strongly on noise and to overlap significantly with the cepstrum distribution of noise. Based on these studies, we suggest a scalar quantity, V, equal to the sum of weighted cepstral coefficients, which is able to classify frames containing speech against noise-like frames. The distributions of V for speech and noise frames are reasonably well separated above SNR = 5 dB, demonstrating the feasibility of robust speech detector based on V.

## 1. INTRODUCTION

Voice Activity Detection (VAD) is one of the traditional issues addressed in speech recognition and telephony applications [1-4]. Depending on the problem in hand, it is either advantageous or absolutely necessary to automatically distinguish between speech and non-speech segments on the early stages of signal processing. In most of the noise suppression and speech enhancement algorithms, for example, one has to estimate correctly noise spectrum. This requires first of all a hard-decision speech pauses detection.

Similarly, in the framework of the acoustic-phonetic approach to speech recognition, when powerful statistical methods have not been yet adopted, a simple speech detection was employed to select only speech frames for a further (phonetic) classification. Various speech detection methods were elaborated then, based first of all on the measurement of short-time energy [1]. Such methods, thanks to their simplicity and effectiveness, are still used in certain applications in telecommunications. In the speech recognition, however, a statistical pattern-classification approach removed the necessity for a preliminary voice detection. In the Hidden Markov Model approach all the irrelevant speech pauses are usually represented by the mixture of gaussian distributions of cepstral coefficients combined into a "silence" state [2]. Silence detection thus became an integral part of the Viterbi classification. The interest to a separate early-stage speech detection was renewed recently in the framework of the auditory-motivated approach to speech recognition. Here, one is interested in an independent probabilistic speech detection or SNR measurement for the purpose of speech labeling prior to recognition [3]. Such label's information can be stored aside and used in the late decision process. Thus, an issue of the proper speech detector emerges.

Both energy-based and gaussian mixture based approaches to speech detection suffer from a lack of robustness and fail to work well in a wide range of SNR, unless some care is taken [2], [4]. Indeed, naïve gaussian distributions are, in a sense, too much specific. They are rather sensitive to noise and, thus, succeed in modeling adequately only a given pre-defined task. On the contrary, short-time energy as a feature is not enough speech-specific: at 0 dB the noise energy becomes comparable to that of speech. Other cues should be added for a robust speech detection, such as LPC error or harmonicity [4]. We suggest below a scalar function of cepstrum coefficients – a measure of cepstral variability – that can be used to classify reliably speech from white noise and to estimate SNR. Such a function provides an independent and complementary to the short-time energy measure of speech presence, thus improving an energy-based detector. On the other hand, it possesses robustness to noise, showing high performance down to 5 dB and reducing the dimensionality of gaussian mixture distributions to the scalar case.

## 2. APPROACH

Speech-like signals are characterized by peculiar patterns in the time-frequency domain. Such patterns can vary from speaker to speaker and from phoneme to phoneme. However, certain features exist that distinguish speech patterns in general from noise, music or other acoustic signals. Revealing such features is the central problem of speech detection that we address.

Apart from transitional and long-term features, all information about the signal is contained in the short-time spectrum, $S(\omega)$. Cepstral representation approximates well the optimal orthogonal decomposition of the short-time spectrum. It maps N (~100-500) Fourier coefficients, or N~20 energy bands, onto ~10 independent cepstrum coefficients that capture most of the relevant local information regarding the signal:

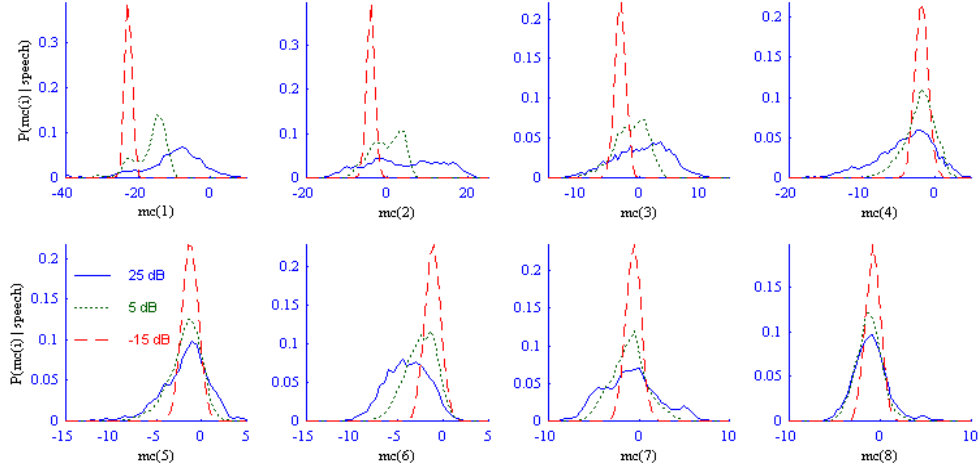

**Figure 1:** Distributions of the first 8 mel-cepstrum coefficients for speech signals. 5000 frames of speech recorded from 1 male speaker in the acoustic room mixed with white gaussian noise at 25 dB (continuous line), 5 dB (dotted line) and –15 dB (dashed line). The distributions of clean speech (continuous line) are characterized in general by the standard deviations larger than that of noise (dashed line). The effect of noise is most prominent on the few lowest cepstral coefficients. The distributions of higher cepstral coefficients of speech tend to become identical to that of noise starting from the 8-th coefficient.

$$c_p = \sum_{k=1}^{N} (\log S_k) \cos\left[\frac{p(k-1/2)\pi}{N}\right] \quad (1)$$

We focus further on the case of mel-cepstrum, where $S_k$ is the output power of each of the N mel-scale bandpass filters, N~20. It can be seen that the mel-cepstrum possesses a build-in robustness, allowing better suppression of insignificant spectral variation in the higher frequency bands than the linear cepstra give.

We assume in the following that the pure speech signal is mixed with an additive gaussian white noise:

$$S(\omega) = S_0(\omega) + \lambda N(\omega), \quad (2)$$

where $\lambda$ is related directly to SNR as $SNR = -20\log_{10}\lambda$ ($S_0(t)$ and $N(t)$ are both normalized to have the standard deviation equal to 1). Unlike convolutive channel noise, additive noise affects cepstrum coefficients in a non-trivial way due to the logarithmic non-linearity in Eq. (1), and cannot be removed by simple mean subtraction methods.

Typical distributions of a few mel-cepstrum coefficients are plotted in Fig 1. Zero's coefficient, $c_0$, plays a role similar to the short-time energy. It is one of the well-known features that distinguishes speech from silence for a clean speech. In the further analysis, however, we rely solely on the information carried by a spectral derivative, $d\log S/d\omega$, omitting $c_0$. Indeed, $c_0$ does not help to discriminate speech from other signals of equal strength, as in the case of strong background noise. The first cepstrum coefficient, $c_1$, reflects in general the overall spectral tilt, low values suggesting frication. It is especially sensitive to transmission channel, speaker and additive noise characteristics. For example, as can be seen from Fig 1, white gaussian noise possesses strong negative values of $c_1$, shifting entirely the corresponding distribution. Higher cepstral coefficients reflect finer spectral details – energy balances between different bands or formants. In particular, $c_2$ quantifies roughly the difference of high and low frequency spectral slopes; $c_3$ approximates the sum of high and low frequency slopes minus the middle-range slope, while $c_4$ can be viewed as a sum of high and low frerquency spectral curvatures, etc. Cepstral coefficients higher than $c_8$ -- $c_{12}$ are influenced by artifacts of numerical analyses (e.g. windowing) and add very little to the accuracy of the method. Thus, a significant for us information is encoded in the first ~8 cepstral coefficients. Following an observation that the range of variation of cepstral coefficients in speech signals is in general higher than that of white noise (Fig 1), we suggest to use the *cepstral variability* as a speech detection feature. The measure of cepstral variability can be quantified by one of the following functions:

$$V_1 = \sum_{i=1}^{D} w_i |c_i|, \quad (3)$$

$$V_2 = \left(\sum_{i=1}^{D} w_i^2 c_i^2\right)^{1/2}, \quad (4)$$

where $w_i$ is so far arbitrary windowing function (note that $V_1$ can take negative values, which differs it qualitatively from $V_2$). As an order of summation we take D=8. Below we refer to $V$ as to *cepstral variability* or *cepstral radius*. The distribution of scores $V$ for speech frames and speech pauses for different SNR are shown in Fig 2. Note that no apriori knowledge about the noise structure is assumed in Eqs (3-4). When the nature of noise is known, Eq (3-4) can be improved further. By encoding information about the noise into the means of its cepstrum distributions, $\bar{c}_{iN}$, we derive:

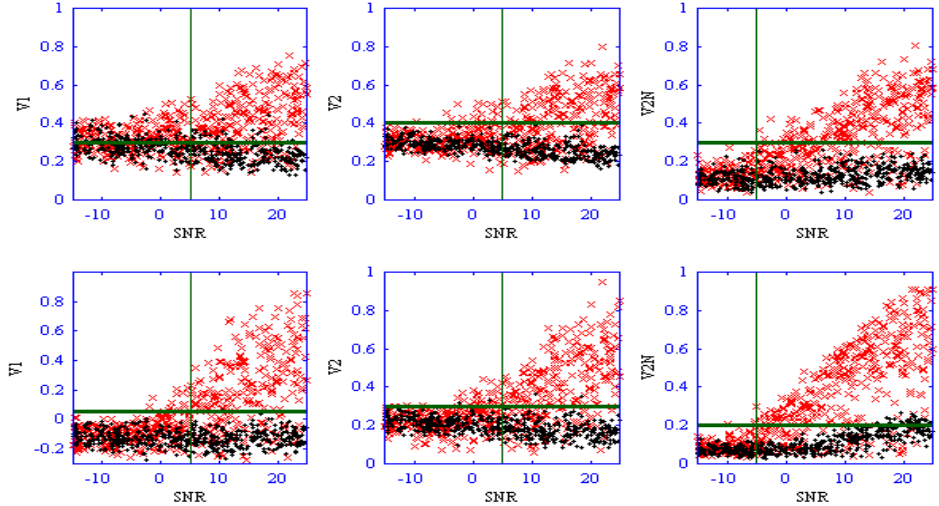

**Figure 2:** Experimentally measured values of cepstral variability scores for 500 speech frames (crosses) and 500 silence frames (dots) mixed with gaussian white noise at different local SNR. The upper row shows distributions computed for the initial window $w_i = i$; the lower row shows the same distributions after window optimization according to Eq (6). Thresholds used in Eq (6) are also shown. In the limit of low SNR (below -10 dB) both speech and silence distributions become identical and collapse to that of white gaussian noise, as expected. The light shift in the distribution of silence frames for high and low SNR hints to the different origin of silence-frame noises as compared to artificially generated white gaussian noise.

$$V_{2N} = \left( \sum_{i=1}^{D} w_i^2 (c_i - \overline{c}_{iN})^2 \right)^{1/2} \quad (5)$$

Low values of $V$ suggest an absence of speech, while high values suggest the presence of speech. Eq (5) can be also regarded as a spectral distortion measure defined in the metric $w_i$, similar to standard LPC and LSF-based spectral distortions. Eqs (3-4) are thus particular distortion measures when one of the signals is 0.

To enhance signal/noise discrimination and to estimate the importance of different cepstral coefficients, we perform optimization of the windowing function $w_i$ using an algorithm based on the signal detection theory. Optimization is performed on a labeled task-specific database. For the data in hand (marked by crosses in Fig 2) we define a fixed threshold $T_{SNR}$ on the SNR axis and an initial threshold $T_V$ on the score axis V. This splits the V-SNR plane into 4 pools. Sensitivity Se is defined as the number of speech scores in the upper right quadrant (high V and high SNR) relative to the number of speech scores for SNR>$T_{SNR}$. Specificity Sp is defined as the number of speech scores in the lower left quadrant (low V and low SNR) relative to the number of speech scores for SNR<$T_{SNR}$. The goal of optimization is to maximize the sum of sensitivity and specificity:

$$\hat{w} = \max_{w} \left[ Se(w, T_{SNR}, T_V) + Sp(w, T_{SNR}, T_V) \right] \quad (6)$$

We initialize the algorithm with $w_i = 1$ or $w_i = i$. Then the threshold $T_V$ is found that maximizes $Se(w_i) + Sp(w_i)$. Finally, Eq (6) is solved for an optimal 8-dimensional vector $\vec{w}$. The resulting value of $\vec{w}$ is normalized so that the highest score V is 1. The new value of $\vec{w}$ is used as an input again, and the process is iterated few times, starting from $T_V$ optimization. The final distributions are shown in the lower row of Fig 2.

Probability of speech presence can be estimated using the Bayes formula. We assign to the pool of speech-like frames all frames containing speech at SNR > 5 dB in Fig 2. The distribution of such frames, $H_s(V)$, is shown in Fig 3 by the dotted line. Similarly, all frames with SNR < -5 dB are assigned to the pool of noise-like frames ($H_n(V)$, plotted in Fig 3 by the dashed-dotted line). On the basis of $H_s(V)$ and $H_n(V)$ we obtain probability of speech given a value of V ($H_s/(H_s + H_n)$, plotted as continuous line in Fig 3). The choice of 2 thresholds at $\pm 5$ dB is made in order to define unambiguously speech-like and noise-like pools in accordance with human perception properties.

Finally, an optimized cepstral variability measure V can be used for a probabilistic SNR estimation. Probability of SNR=T, derived from the distributions in Fig 2 for different $V_2$ values is shown in Fig 4.

## 3. EXPERIMENTAL RESULTS

5000 frames of clean speech, 128 ms each, were recorded at 8kHz sampling rate from one speaker in the acoustic room where effects of external noise and echo are absent. The same amount of non-speech frames was selected as speech pauses of the same recording, containing breathing and computer noise. All frames were mixed with the white gaussian noise at different local frame SNR, Eq (2), ranging from –15 to 25 dB, pre-emphasized and Hamming-windowed. A bank of 16 mel-scale Hanning-window filters covering the range from 100 to 3500 Hz was applied to compute mel-cepstrum (Fig 1). 4000 frames contributed to tune the measure V by optimizing $w_i$ in Eqs (3-5) using Eq (6) and the Nelder-Mead type simplex search method. The best results of optimization were achieved for $V_{2N}$ (Se+Sp = 1.77, $\vec{w}$ = [.7, .8, .8, 1., .4, .6, .8, .1]); then follows $V_1$ (Se+Sp = 1.64, $\vec{w}$ = [-.4, .2, .3, .3, -.2, 1., .3, -.1]), and $V_2$ (Se+Sp=1.58, $\vec{w}$ = [.01, .6, .4, .7, .5, 1., .6, .7]). In all cases few iterations were performed to arrive at the optimal window $\vec{w}$. The threshold $T_{SNR}$ =5 dB was set for $V_1$ and $V_2$ optimization, while $T_{SNR}$ = -5 dB for $V_{2N}$. The optimized functions were used to build a probabilistic speech detector and set a binary decision threshold (P(speech|V)=0.5 in Fig 3). The remaining 1000 frames, classified *a priori* as speech according to the threshold at SNR=5 dB, and non-speech frames were used to test the performance. The results of the binary decision process are collected in Table 1. The results of Nelder-Mead type optimization show very little dependence on the choice of initial $w_i$. For $V_2$, however, the linear window outperforms the uniform window yet before the optimization. This is in part due to the fact that the white gaussian noise possesses strong negative values of $c_1$. When contributing to $V_2$ on the equal footing with higher cepstral coefficients, large $c_1$ tend to obscure the difference between speech and non-speech frames at low SNR. To improve the separation between speech and noise, one has to suppress $c_1$. This is what indeed happens in the process of optimization. Finally, all the steps of the algorithm were repeated for the lynx noise from NOISEX database. Although the spectral shape of lynx noise differs drastically from the white gaussian one, the results were quite similar.

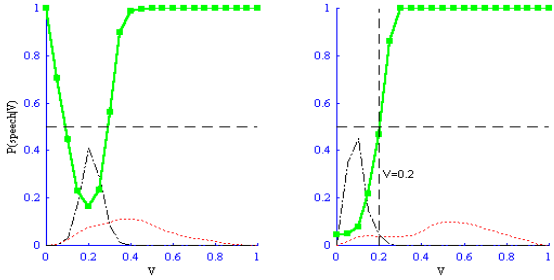

**Figure 3:** Bayesian probability of speech presence, P(speech|V), as a function of cepstral variability score V (continuous line), obtained from distributions of V for speech (dotted line) and noise (dashed-dotted line). Hard decision thresholds at P=0.5 are shown by dashed lines. Left plot corresponds to $V_2$ case, while the right one to $V_{2N}$.

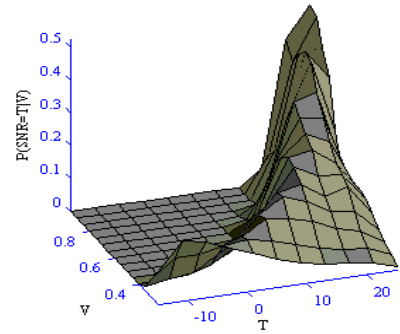

**Figure 4:** Probabilistic estimation of SNR on the basis of measured values of optimized cepstral variability $V_2$.

|  | $V_1$ | $V_2$ | $V_{2N}$ |
|---|---|---|---|
| **speech** | 77% | 74% | 88% |
| **non-speech** | 95% | 94% | 98% |

**Table 1:** Speech detection rates for speech and rejection rates for non-speech frames for 3 different detectors.

## 4. CONCLUSION

We proposed a cepstral variability feature, V, and demonstrated that it can be useful for a robust speech-noise discrimination and SNR estimation in noisy environments. $V_{2N}$ was shown to perform the best, followed by $V_1$ and $V_2$. Since V is complementary to the short-time energy, it can improve the reliability of VAD when used either as a separate additional parameter, or in a particular combination with the energy: $\alpha \log E + \beta \log V$ or $\log(aE + bV)$. Cepstral variability can be viewed as a robust scalar analogue of the traditional HMM "silence" state, or as a generalized spectral distortion. Experiments were performed for the additive white gaussian noise and the lynx noise from NOISEX, and we expect that the results must hold also for any stationary white (wide-band) noises. Indeed, due to the random nature of such noises, cancellations among different parts of the spectrum will force the cepstral variability to stay close to zero. Our results refer to the off-line mode of work with a stationary noise, but can be extended to the case of an on-line optimization with a noise adaptation in Eq (5).

**ACKNOWLEDGEMENTS:** This work was supported by the European Union under the RESPITE project.

## REFERENCES

[1] S.VanGerven and F. Xie, "A Comparative Study Of Speech Detection Methods," *Eurospeech-97*, p.1095.
[2] B.McKinley and G.Whipple "Model Based Speech Pause Detection," Proc. ICASSP-97, p.1179
[3] F.Berthommier and H.Glotin, "A Measure Of Speech And Pitch Reliability From Voicing," Proc CASA workshop, IJCAI, Stockholm (1999) p.61-70.
[4] H.Kobatake, K.Tawa and A.Ishida, "Speech/Nonspeech Discrimination For Speech Recognition System Under Real Life Noise Environments," Proc. ICASSP-89, p.365.